\pdfoutput=1

\documentclass[11pt]{article}

\usepackage[preprint]{acl}

\usepackage{times}
\usepackage{bm}
\usepackage{latexsym}
\usepackage{amsmath} 
\usepackage{multirow}
\usepackage{booktabs}
\usepackage{array}
\usepackage{amssymb}
\usepackage{pifont}

\usepackage{url}
\usepackage[T1]{fontenc}
\usepackage[utf8]{inputenc}
\usepackage{microtype}
\usepackage{inconsolata}
\usepackage{graphicx}

\title{Dolphin: Document Image Parsing via Heterogeneous Anchor Prompting}

\author{Hao Feng$^*$, 
Shu Wei$^*$, 
Xiang Fei\thanks{The first four authors contributed equally to this work.},
Wei Shi$^{*\dag}$,
Yingdong Han,
Lei Liao,
 \\ 
\textbf{
Jinghui Lu,
Binghong Wu, 
Qi Liu, 
Chunhui Lin, 
Jingqun Tang,
Hao Liu,
Can Huang\thanks{Corresponding author}} \\
\\
\textbf{ByteDance}}

\begin{document}
\maketitle
\begin{abstract}
Document image parsing is challenging due to its complexly intertwined elements such as text paragraphs, figures, formulas, and tables. 
Current approaches either assemble specialized expert models or directly generate page-level content autoregressively, facing integration overhead, efficiency bottlenecks, and layout structure degradation despite their decent performance. 
To address these limitations, we present \textit{Dolphin} (\textit{\textbf{Do}cument Image \textbf{P}arsing via \textbf{H}eterogeneous Anchor Prompt\textbf{in}g}), a novel multimodal document image parsing model following an analyze-then-parse paradigm. 
In the first stage, Dolphin generates a sequence of layout elements in reading order. 
These heterogeneous elements, serving as anchors and coupled with task-specific prompts, are fed back to Dolphin for parallel content parsing in the second stage. 
To train Dolphin, we construct a large-scale dataset of over 30 million samples, covering multi-granularity parsing tasks. 
Through comprehensive evaluations on both prevalent benchmarks and self-constructed ones, Dolphin achieves state-of-the-art performance across diverse page-level and element-level settings, while ensuring superior efficiency through its lightweight architecture and parallel parsing mechanism.
The code and pre-trained models are publicly available at 
\href{https://github.com/ByteDance/Dolphin}{{\tt https://github.com/ByteDance/Dolphin}}
\end{abstract}

\begin{figure}[t]
    \centering
  \includegraphics[width=1\columnwidth]{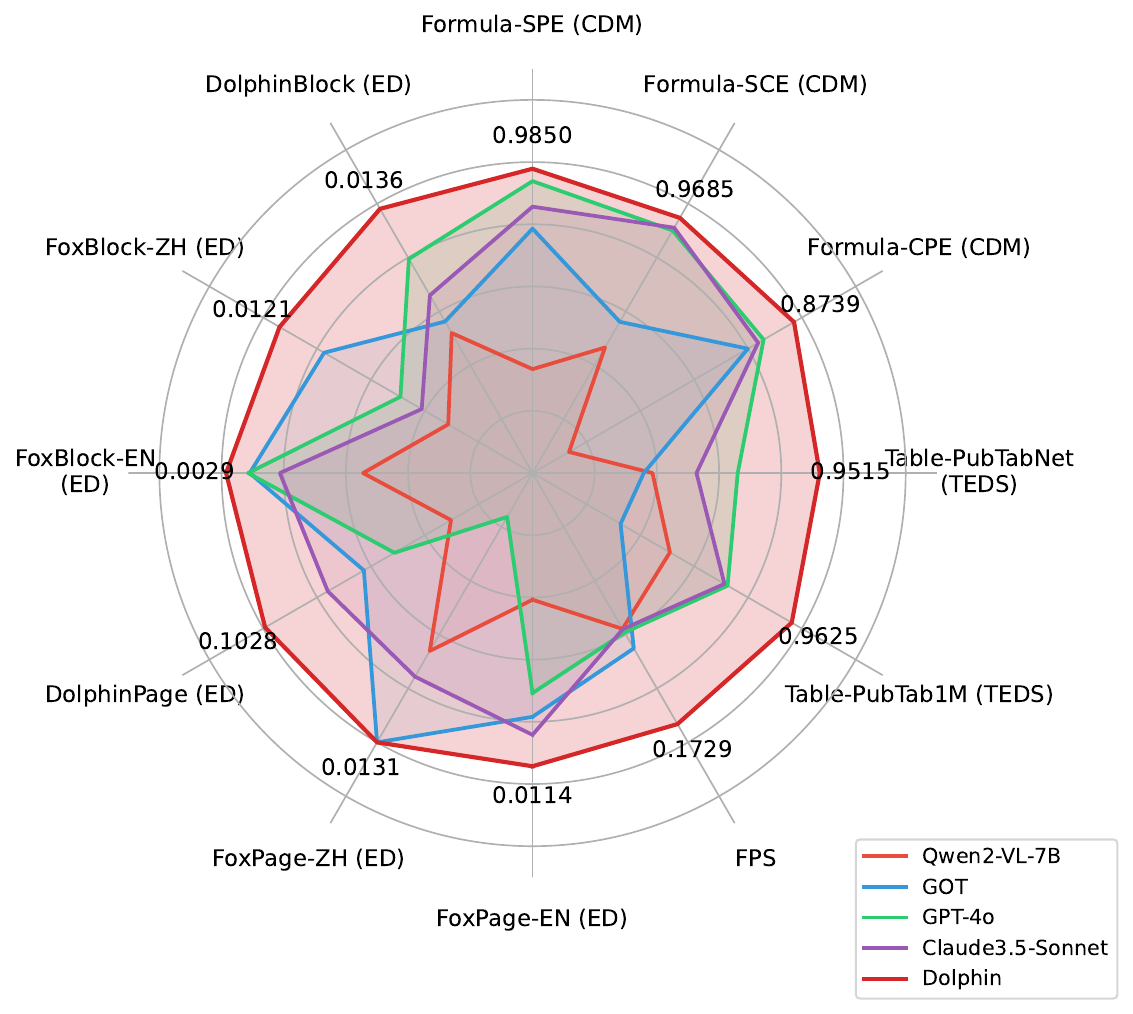}
    \caption{Comparison of Dolphin with advanced VLMs across benchmarks: page-level parsing (plain and complex documents), element-level parsing (text paragraph, table, and formula), and running efficiency (FPS). The outer area represents better performance. Dolphin exhibits the best performance in most evaluations.}
  \label{fig:polar}
\end{figure}

\section{Introduction}
Document image parsing~\citep{blechernougat} aims to extract structured content from images containing intertwined elements such as text paragraphs, figures, tables, and formulas.
As a foundational capability for downstream content analysis~\citep{wang2023docllm}, it bridges the gap between visual content and machine-readable formats. With the exponential growth of digital documents across domains like academic papers, business reports, and technical documentation, robust document parsing capabilities have become increasingly critical.

Current document image parsing solutions have evolved along two distinct trajectories.
The first one~\citep{wang2024mineru} integrates specialized models for different OCR~\citep{tang2022few,zhao2024multi,tang2022youcan} tasks (\textit{e.g.}, layout detection, reading order prediction, and recognition for textlines, formulas, and tables). These solutions demonstrate strong performance through dedicated expertise, but require independent optimization of each model and face coordination challenges across components. To address these challenges, recent works leverage general or expert vision-language models (VLMs)~\citep{liu2024visual} to directly generate page-level content autoregressively, benefiting from end-to-end training and effective multimodal feature fusion. These methods~\citep{blechernougat,kim2022ocr,wei2024general} show impressive results in capturing page-level semantics. However, they also encounter layout structure degradation and efficiency bottlenecks when parsing long documents with complex layouts.

To synergize the advantages of both approaches while addressing their limitations, we present \textit{Dolphin} (\textit{\textbf{Do}cument Image \textbf{P}arsing via \textbf{H}eterogeneous Anchor Prompt\textbf{in}g}), a novel vision-language model following an \textit{analyze-then-parse} paradigm. Rather than relying on multiple expert models or purely autoregressive generation, Dolphin decomposes document parsing into two strategic stages. In the first stage, Dolphin performs comprehensive page-level layout analysis by generating an element sequence in natural reading order, while preserving rich structural relationships (\textit{e.g.}, figure-caption pairs, table-caption associations, and section title-paragraph hierarchies). These analyzed elements then serve as anchors for the second stage, where element-specific prompts enable efficient parallel parsing of multiple elements. The focused context within each element allows the vision-language model to effectively recognize the document contents.

To train Dolphin on different granularities of tasks, we construct a large-scale dataset of 30 million samples containing both page-level documents and element-level blocks. Notably, Dolphin's element-decoupled parsing strategy offers unique advantages in data collection, as acquiring isolated element images (\textit{e.g.}, tables, formulas) and their annotations is more feasible than collecting full document pages with diverse elements.

Comprehensive evaluations are conducted on prevalent benchmarks and self-constructed ones. The results show that Dolphin achieves state-of-the-art performance across diverse page-level and element-level parsing tasks (Figure~\ref{fig:polar}). Moreover, benefiting from its lightweight architecture and parallel element parsing mechanism, Dolphin exhibits considerable advantages in running efficiency.

\begin{figure*}[t]
  \centering
  \includegraphics[width=1.98\columnwidth]{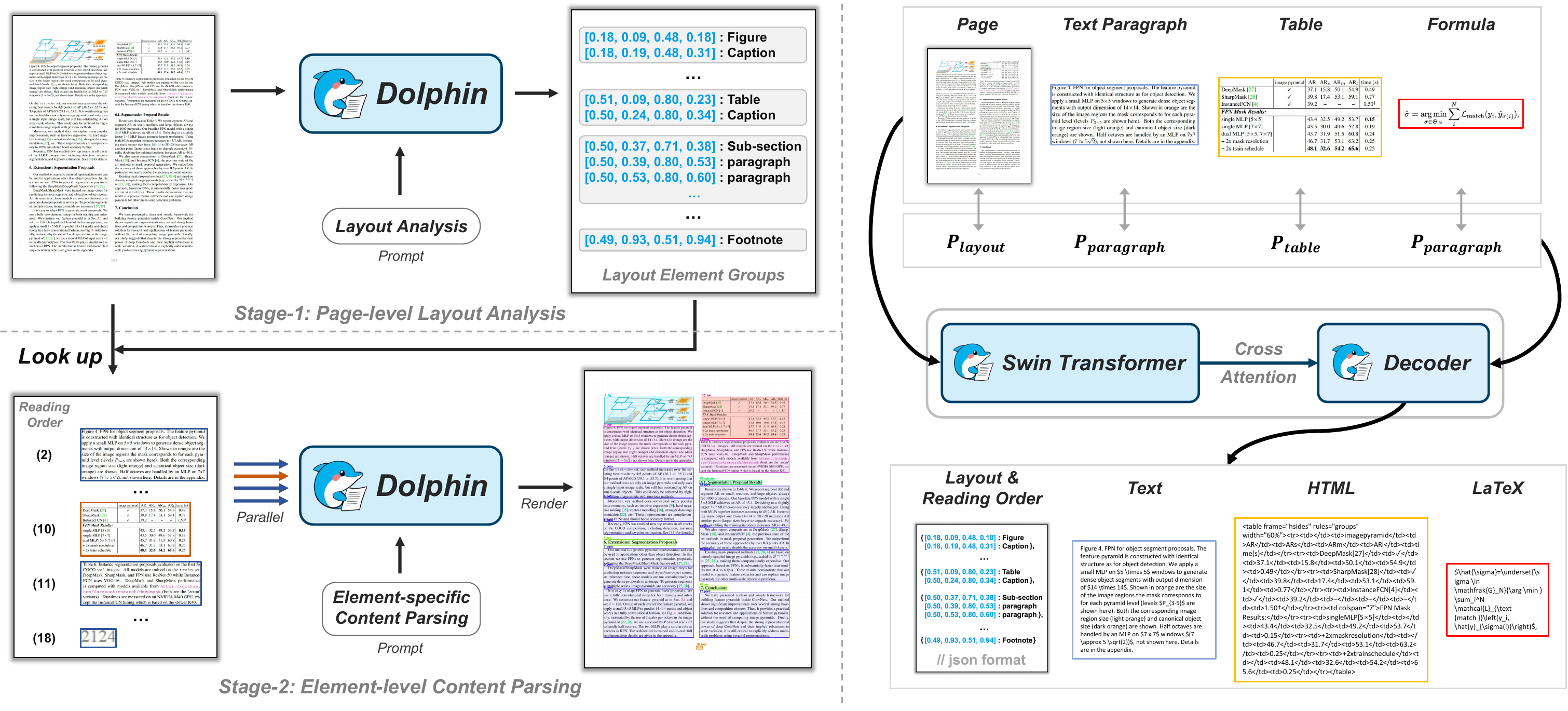}
\caption{Overview of Dolphin's two-stage document image parsing paradigm. \textbf{Left}: The pipeline consists of Stage 1 for page-level layout analysis that generates structured layout sequences in reading order, and Stage 2 for element-level content parsing. \textbf{Right}: Examples of input-output pairs, including page-level layout analysis and element-level content parsing for text paragraphs, tables, and formulas. ``$P_*$" denotes different prompts.}
      \label{fig:framework}
\end{figure*}

\section{Related Work}
Document image parsing enables robust content extraction from rendered document images without relying on source file formats or parsing libraries (\textit{e.g.}, PyMuPDF).
Existing solutions can be categorized into two streams: integration-based methods that assemble multiple expert models in a pipeline, and end-to-end approaches that leverage vision-language models to directly generate structured results via autoregressive decoding.

\subsection{Integration-based Document Parsing}
Traditional document parsing solutions rely on integrating multiple specialized models in a multi-stage pipeline~\citep{xu2020layoutlm,herzig2020tapas,zhang2017watch}. These approaches typically begin with layout detection to identify different types of elements (\textit{e.g.}, tables, formulas), followed by dedicated recognizers for each element type. Recent commercial and academic solutions such as Mathpix\footnote{https://mathpix.com/pdf-conversion/}, TextIn\footnote{https://www.textin.ai/}, 
and MinerU~\citep{wang2024mineru} follow this integration-based paradigm. Notably, MinerU advances this direction by introducing sophisticated content filtering and segmentation strategies. These methods demonstrate strong performance through specialized expertise and have shown significant potential in high-precision content extraction. However, they face challenges in system complexity, cross-model coordination, and limited understanding of complex document layouts when compared to end-to-end approaches.

\subsection{Autoregressive Document Parsing}
Recent advances in vision-language models have enabled a new paradigm of end-to-end document image parsing, categorized into two streams.

\smallskip
\textbf{General VLMs.} With the rapid development of large vision-language models, researchers have begun exploring the application of general-purpose VLMs~\citep{liu2024visual} to document parsing tasks. Models such as GPT-4V~\citep{yang2023dawn}, Claude-series\footnote{https://www.anthropic.com/news/claude-3-5-sonnet}, Gemini-series~\citep{team2024gemini}, QwenVL-series~\citep{wang2024qwen2,bai2025qwen2}, MiniCPM-series~\citep{yao2024minicpm}, InternVL-series~\citep{chen2024internvl}, DeepSeek-VL2~\citep{wu2024deepseek}, and Step-1V demonstrate promising results in document understanding without task-specific training. These models benefit from large-scale pre-training on diverse visual data, exhibiting strong zero-shot capabilities. However, they frequently face challenges in processing efficiency, specialized element recognition, and layout structure preservation, particularly when processing long documents with complex layouts.

\smallskip
\textbf{Expert VLMs.} These models are specifically designed and trained for document parsing or understanding tasks. Nougat~\citep{blechernougat} pioneered this direction by introducing an encoder-decoder model that converts documents into markup language. GOT~\citep{wei2024general} presented an innovative unified model that processes various document elements. Other representative work such as Donut~\citep{kim2022ocr}, LayoutLM-series~\citep{xu2020layoutlm,xu2020layoutlmv2,huang2022layoutlmv3}, UDOP~\citep{tang2023unifying}, Wukong-Reader~\citep{bai2023wukong}, KOSMOS-series~\citep{lv2023kosmos,peng2023kosmos}, UniDoc~\citep{feng2023unidoc}, UReader~\citep{ye2023ureader}, DocPedia~\citep{feng2024docpedia}, TGDoc~\citep{wang2023towards}, Vary~\citep{wei2024vary}, Fox~\citep{liu2024focus}, Monkey-series~\citep{li2024monkey,liu2024textmonkey}, TabPedia~\citep{zhao2024tabpedia}, TextSquare~\citep{tang2024textsquare}, DocFusion~\citep{chai2024docfusion}, TextHawk-series~\citep{yu2024texthawk,yu2024texthawk2}, mPLUG-DocOwl-series~\citep{ye2023mplug,hu2024mplug}, SmolDocling~\citep{nassar2025smoldocling}, PlatPus~\citep{wang2024platypus}, olmOCR~\citep{poznanski2025olmocr}, Ocean-OCR~\citep{chen2025ocean}, and Mistral-OCR\footnote{https://mistral.ai/fr/news/mistral-ocr} have been proposed. Despite their impressive performance, these expert VLMs face similar challenges as general VLMs.

\section{Approach}
In this section, we present our Dolphin in detail. We first provide an overview of our analyze-then-parse paradigm, followed by detailed descriptions of the page-level layout analysis stage and element-level content parsing stage.

\subsection{Overview}
Dolphin follows an analyze-then-parse paradigm built upon an encoder-decoder transformer architecture. As shown in Figure~\ref{fig:framework} (left), given an input document image $I$, the first stage performs page-level layout analysis to extract elements in reading order. These elements then serve as anchors for the second stage, where type-specific prompts guide parallel parsing of individual elements. The core of both stages is a unified vision-language model, which shares the same parameters but operates on different input granularities with distinct prompting strategies, as presented in Figure~\ref{fig:framework} (right).

\begin{table*}[t]
\setlength{\tabcolsep}{0.98mm}
\small
\centering
\begin{tabular}{llccccccc}
\hline
\multirow{2}{*}{\textbf{Category}} & \multirow{2}{*}{\textbf{Method}} & \multirow{2}{*}{\textbf{Model Size}} & \multicolumn{2}{c}{\textbf{Plain Doc (ED $\downarrow$)}} & \textbf{Complex Doc (ED $\downarrow$)} & \multirow{2}{*}{\textbf{Avg. ED}} & \multirow{2}{*}{\textbf{FPS} $\uparrow$} \\
\cmidrule(lr){4-5} \cmidrule(lr){6-6}
& & & Fox-Page-EN & Fox-Page-ZH & Dolphin-Page & & \\
\hline
\multirow{2}{*}{Integration-based} & MinerU & 1.2B & 0.0685 & 0.0702 & 0.2770 & 0.1732 & 0.0350 \\
& Mathpix & - & 0.0126 & 0.0412 & 0.1586 & 0.0924 & 0.0944 \\
\hline
\multirow{8}{*}{Expert VLMs} & Nougat & 250M & 0.1036 & 0.9918 & 0.7037 & 0.6131 & 0.0673 \\
& Kosmos-2.5 & 1.3B & 0.0256 & 0.2932 & 0.3864 & 0.2691 & 0.0841 \\
& Vary & 7B & 0.092$^*$ & 0.113$^*$ & - & - & - \\
& Fox & 1.8B & 0.046$^*$ & 0.061$^*$  & - & - & - \\
& GOT & 580M & 0.035$^*$ & 0.038$^*$  & 0.2459 & 0.1411 & 0.0604 \\
& olmOCR & 7B & 0.0235 & 0.0366 & 0.2000 & 0.1148 & 0.0427 \\
& SmolDocling & 256M & 0.0221 & 0.7046 & 0.5632 & 0.4636 & 0.0140 \\
& Mistral-OCR & - & 0.0138 & \underline{0.0252} & \underline{0.1283} & \underline{0.0737} & 0.0996 \\
\hline
\multirow{12}{*}{General VLMs}
& InternVL-2.5 & 8B & 0.3000 & 0.4546 & 0.4346 & 0.4037 & 0.0444 \\
& InternVL-3 & 8B & 0.1139 & 0.1472 & 0.2883 & 0.2089 & 0.0431 \\
& MiniCPM-o 2.6 & 8B & 0.1590 & 0.2983 & 0.3517 & 0.2882 & 0.0494 \\
& GLM4v-plus & 9B  & 0.0814 & 0.1561 & 0.3797 & 0.2481 & 0.0427 \\
& Gemini-1.5 pro & - & 0.0996 & 0.0529 & 0.1920 & 0.1348 & 0.0376 \\
& Gemini-2.5 pro & - & 0.0560 & 0.0396 & 0.2382 & 0.1432 & 0.0231 \\
& Claude3.5-Sonnet & - & 0.0316 & 0.1327 & 0.1923 & 0.1358 & 0.0320 \\
& GPT-4o-202408 & - & 0.0585 & 0.3580 & 0.2907 & 0.2453 & 0.0368 \\
& GPT-41-250414 & - & 0.0489 & 0.2549 & 0.2805 & 0.2133 & 0.0337 \\
& Step-1v-8k & - & 0.0248 & 0.0401 & 0.2134 & 0.1227 & 0.0417 \\
& Qwen2-VL & 7B & 0.1236 & 0.1615 & 0.3686 & 0.2550 & 0.0315 \\
& Qwen2.5-VL & 7B & \underline{0.0135} & 0.0270 & 0.2025 & 0.1112 & 0.0343 \\
\hline
\multirow{1}{*}{Ours} & \textbf{Dolphin} & 322M & \textbf{0.0114} & \textbf{0.0131} & \textbf{0.1028} & \textbf{0.0575} & \textbf{0.1729} \\
\hline
\end{tabular}
\caption{\label{tab:benchmark-results}
Performance comparison of \textbf{page-level document parsing}. ``Plain Doc'' represents documents containing only text content, while ``Complex Doc'' includes documents with mixed elements (tables, formulas, and figures). Arrow ``$\uparrow$/$\downarrow$'' indicate whether higher/lower values are better. Results marked with ``$^*$'' are reported by GOT. \textbf{Boldface} indicates the best performance and \underline{underlined} values denote the second-best.
}
\end{table*}

\subsection{Page-level Layout Analysis}
This stage aims to identify the layout elements and their reading order through the following steps: 

\smallskip
\textbf{Page Image Encoding.} We employ Swin Transformer~\citep{liu2021swin} as our visual encoder, which takes the page image $I$ as input and outputs a sequence of visual embeddings $z \in \mathbb{R}^{d\times N}$, where $d$ is the embedding dimension and $N$ is the number of image patches. The hierarchical design of Swin enables capturing both global layout patterns and local textual details. Note that the input image is resized and padded to a fixed size of $H \times W$ while preserving its aspect ratio to avoid text distortion.

\begin{figure*}[t]
    \centering
    \includegraphics[width=1.82\columnwidth]{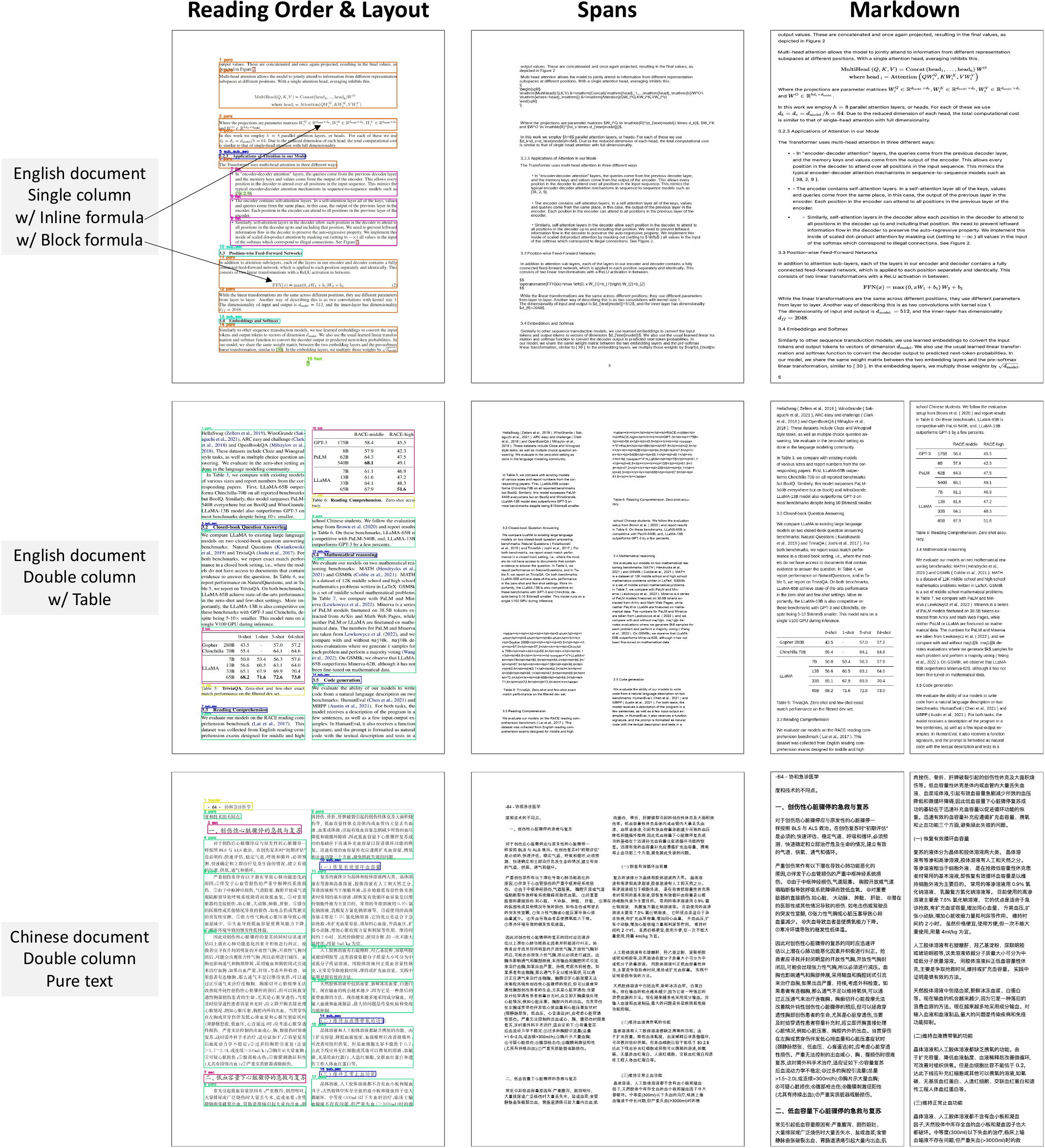}
    \caption{Visualization of Dolphin's \textbf{page-level} parsing results. \textbf{Left}: Layout analysis form Stage 1 with predicted element boundaries and reading order. \textbf{Middle}: Element-specific parsing outputs from Stage 2. \textbf{Right}: Final rendered document in markdown format. More cases are shown in the supplementary material.}
  \label{fig:page_visualization}
\end{figure*}

\smallskip
\textbf{Layout Sequence Generation.} Taking the layout analysis prompt $P_{layout}$ as a guide, the decoder attends to the encoded visual features through the cross-attention mechanism~\citep{vaswani2017attention}. We adopt mBart~\citep{lewis2019bart} as the decoder. With the prompt "\textit{Parse the reading order of this document.}", the model identifies and arranges document elements sequentially, while preserving structural relationships (\textit{e.g.}, figure-caption pairs, table-caption associations, and section title-paragraph hierarchies). As shown in Figure~\ref{fig:framework}, it generates a sequence of layout elements $L = \{l_1, l_2, ..., l_n\}$, where element $l_i$ specifies its type (\textit{e.g.}, \textit{figure, caption, table, paragraph}) and bounding box. This structured layout sequence provides anchors for the subsequent element-level parsing stage.

\subsection{Element-level Content Parsing}
The second stage leverages the analyzed layout elements as anchors for parallel element parsing. This design marks a key departure from purely autoregressive approaches, enabling efficient processing while maintaining element-specific expertise. We achieve this through two steps:

\textbf{Element Image Encoding.} For each layout element $l_i$ identified in the first stage, we crop its corresponding region from the original image to create a local view $I_i$. These local views are encoded in parallel using the same Swin Transformer, producing element-specific visual features.

\textbf{Parallel Content Parsing.} With the encoded element features, we employ type-specific prompts to guide the parsing of different elements. As shown in Figure~\ref{fig:framework} (right), tables employ dedicated prompts $P_{table}$ to parse their HTML format, while formulas share the same prompt $P_{paragraph}$ as text paragraphs since they frequently appear both inline and in display mode within paragraph context, despite their LaTeX markup format. Given the visual feature of the local view $I_i$ and its corresponding prompt $p_i$, the decoder generates the parsed content in parallel. This parallel processing strategy, combined with element-specific prompting, ensures computational efficiency while maintaining accurate content recognition.

\section{Dataset}
To enable comprehensive training and evaluation, we construct large-scale datasets spanning multiple document granularities and parsing tasks.

\subsection{Training}
For training, we collect over 30 million samples covering both page-level documents and element-level components. A comprehensive breakdown of our training dataset, including data sources, granularities, and task, is shown in Table~\ref{tab:dataset}. In the following, we describe the preparation and collection of data for different training objectives.

\textbf{Mixed Documents.} We collect 0.12M documents from diverse sources, including educational materials (exam papers and textbooks), publications (magazines and newspapers), and business documents (presentations and industry reports). All documents are annotated with element-level boundaries and their reading order, enabling training for both layout analysis and order prediction.

\setlength{\tabcolsep}{0.9mm}
\begin{table}[t]
  \small
  \centering
  \begin{tabular}{lccc}
    \hline
    \textbf{Source} & \textbf{Granularity} & \textbf{\#Samples} & \textbf{Task Types} \\
    \hline
    Mixed Documents & Page & 0.12M & Layout \\
    HTML & Page & 4.37M & Parsing \\
    LaTeX & Page & 0.5M & Parsing \\
    Markdown & Page & 0.71M & Parsing \\
    Table & Element & 1.57M & Parsing \\
    Formula & Element & 23M & Parsing \\
    \hline
    Total & - & 30.27M & - \\
    \hline
  \end{tabular}
  \caption{\label{tab:dataset}
        Overview of our training data. Note that page-level documents are also decomposed into individual elements for element-specific training.
    }
\end{table}

\begin{table*}[t]
\setlength{\tabcolsep}{1.05mm}
\small
\centering
\begin{tabular}{llcccccccc}
\hline
\multirow{4}{*}{\textbf{Category}} & \multirow{4}{*}{\textbf{Method}} & \multicolumn{3}{c}{\textbf{Text Paragraph (ED $\downarrow$)}} & \multicolumn{3}{c}{\textbf{Formula (CDM $\uparrow$)}} & \multicolumn{2}{c}{\textbf{Table (TEDS $\uparrow$)}} \\
\cmidrule(lr){3-5} \cmidrule(lr){6-8} \cmidrule(lr){9-10}
& & \multicolumn{2}{c}{Fox-Block} & \multirow{2}{*}{Dolphin-Block} & \multirow{2}{*}{SPE} & \multirow{2}{*}{SCE} & \multirow{2}{*}{CPE} & \multirow{2}{*}{PubTabNet} & \multirow{2}{*}{PubTab1M} \\
\cmidrule(lr){3-4}
& & EN & ZH \\
\hline
\multirow{3}{*}{Expert Models} & UnimerNet-base & - & - & - & \textbf{0.9914}$^*$ & 0.94$^*$ & 0.9595$^*$ & - & - \\
& Mathpix & - & - & - & 0.9729$^*$ & 0.9318$^*$ & \textbf{0.9671}$^*$  & - & - \\
& Pix2tex & - & - & - & 0.9619$^*$ & 0.2453$^*$ & 0.6489$^*$ & - & - \\
\hline
\multirow{2}{*}{Expert VLMs}
& TabPedia  & - & - & - & - & - & - & \textbf{0.9541} & 0.9511 \\
& GOT & 0.0181 & 0.0452 & 0.0931 & 0.8501 & 0.7369 & 0.7197 & 0.3684 & 0.3269 \\
\hline
\multirow{7}{*}{General VLMs} 
& GLM-4v-plus & 0.0170 & 0.0400 & 0.1786 & 0.9651 & 0.9585 & 0.7055 & 0.5462  & 0.6018  \\
& Qwen2-VL-7B & 0.0910 & 0.1374 & 0.1012 & 0.5339 & 0.6797 & 0.1220 & 0.3973 & 0.5101  \\
& Qwen2.5-VL-7B & 0.0803 & 0.0301 & 0.0712 & 0.9486 & 0.9484 & 0.8309 & 0.6169 & 0.6462  \\
& Gemini-1.5 pro & 0.0108 &  0.0461 & 0.0857 & 0.9572 & 0.9469 &  0.7171  & 0.7571 & 0.7776  \\
& Claude3.5-Sonnet & 0.0375 & 0.1177 & 0.0746 & 0.8995 & 0.9464 & 0.7543 & 0.5431 &  0.7127  \\
& GPT-4o-202408 & 0.0170 & 0.1019 & 0.0489 & 0.9570 & 0.9402 & 0.7722 &  0.6692 & 0.7243  \\
& Step-1v-8k & 0.0098 & 0.0175 & 0.0252 & 0.9526 & 0.9336 & 0.7519 & 0.6808 & 0.6588  \\
\hline
\multirow{1}{*}{Ours} & \textbf{Dolphin} & \textbf{0.0029} & \textbf{0.0121} & \textbf{0.0136} & 0.9850 & \textbf{0.9685} & 0.8739 & 0.9515 & \textbf{0.9625} \\
\hline
\end{tabular}
\caption{\label{tab:benchmark-results-block}
Performance comparison of \textbf{element-level parsing} across text paragraphs, formulas, and tables. Arrows ``$\uparrow$/$\downarrow$'' indicate whether higher/lower values are better. Results marked with ``$^*$'' are reported by UnimerNet.
}
\end{table*}

\textbf{HTML.} For documents from the HTML source, we utilize dumps from Chinese and English Wikipedia articles to generate synthetic training data through web rendering~\citep{kim2023web}. We process HTML content by adding span tags for character-level annotation, and apply random font selection to enhance visual diversity. Through this pipeline, we generate 4.37M page-level samples with comprehensive bounding box annotations at character, word, line and paragraph levels.

\textbf{LaTeX.} We collect 0.5M documents from the arXiv database and process them using LaTeX Rainbow~\citep{duanlatex}, a specialized rendering framework that preserves document hierarchical structure. This tool renders different element (\textit{e.g.}, formulas, figures) with distinct colors while maintaining the reading order. The rendered documents are then automatically parsed to extract element types, hierarchical relationships, and spatial locations at block, line, and word levels.

\textbf{Markdown.} We collect 0.71M markdown documents from GitHub pages and process them using Pandoc~\citep{macfarlane2013pandoc} for PDF rendering with several customized templates. Through PyMuPDF-based parsing and content alignment with source markdown, we obtain hierarchical text annotations at paragraph, line, and word levels, as well as some specific element types like tables. Furthermore, we render the formula in different colors and find all formula blocks based on pixel matching.

\textbf{Tables.} For table parsing, we utilize PubTabNet~\citep{zhong2020image} and PubTab1M~\citep{smock2022pubtables}, two large-scale datasets of tables extracted from scientific publications. PubTabNet contains 568K tables with HTML annotations, while PubTab1M provides 1M tables with more fine-grained structure annotations.

\textbf{Formulas.} We collect 23M formula expressions in LaTeX format from arXiv sources, including in-line formulas, single-line formulas, and multi-line formulas. The expressions are then rendered formula images using the XeTeX tool. Various backgrounds and fonts are used in the rendering process to enhance the richness of the images.

\subsection{Evaluation}
The evaluation is conducted at both the page and the element levels. At the page level, we evaluate the models on two distinct benchmarks: Fox-Page~\citep{liu2024focus}, which consists of pure text documents, and our newly constructed Dolphin-Page containing complex documents with interleaved figures, tables, and mathematical formulas. At the element level, we assess the fine-grained parsing capabilities for text-paragraph, formulas, and tables through the public test sets.

\smallskip
\textbf{Page-level Evaluation:}

\textbf{\textit{(a) Fox-Page.}} Fox-Page is a bilingual benchmark containing 212 document pages (112 in English and 100 in Chinese) including both single-column and multi-column formats. Each page contains over 1,000 words, making it a challenging testbed for document image parsing. 

\textbf{\textit{(b) Dolphin-Page.}} Our Dolphin-Page is a bilingual benchmark of 210 document pages designed for complex document parsing. It consists of 111 pure text documents and 99 challenging samples with interleaved tables, mathematical formulas, and figures in both single-column and multi-column layouts. All documents are manually annotated with precise transcriptions following the natural reading order, making it a rigorous testbed for evaluating document parsing capabilities.

\begin{figure*}[t]
    \centering
    \includegraphics[width=1.98\columnwidth]{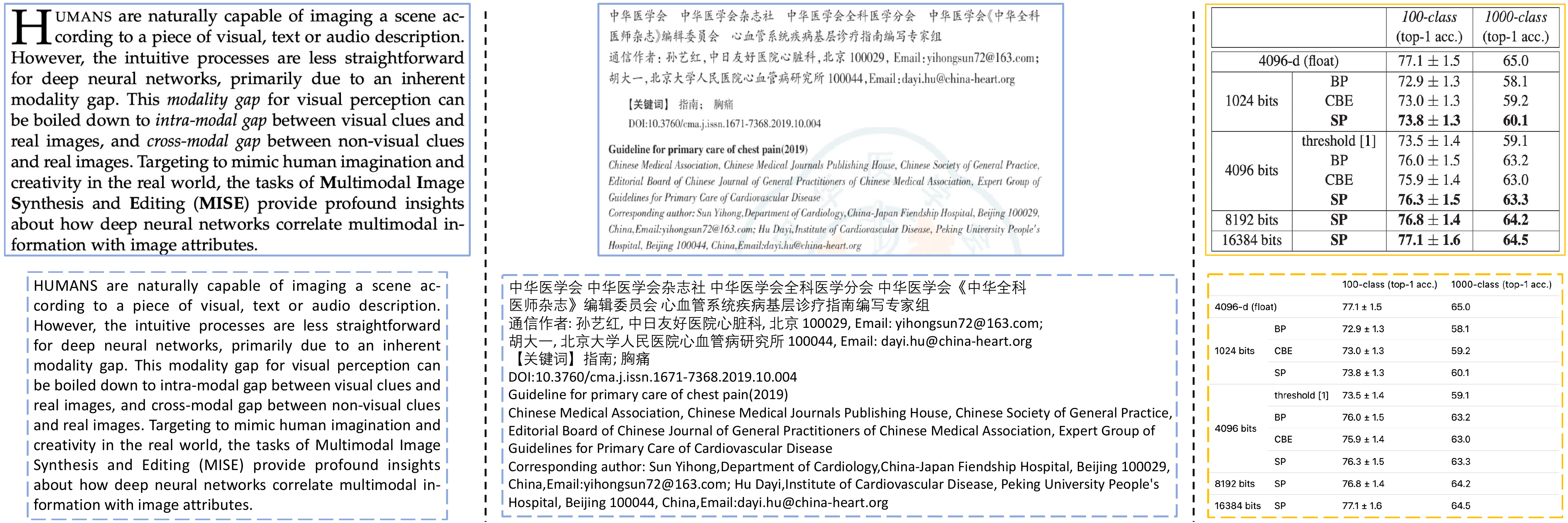}
    \caption{Demonstration of Dolphin's \textbf{element-level} parsing across diverse scenarios. Input images are shown in the top row, with corresponding recognition results in the bottom row. \textbf{Left}: Text paragraph parsing in complex layouts. \textbf{Middle}: Bilingual text paragraph recognition. \textbf{Right}: Complex table parsing (rendered results shown).}
   \label{fig:network}
\end{figure*}

\smallskip
\textbf{Element-level Evaluation:}

\textbf{\textit{(a) Text Paragraph}.} For pure text recognition evaluation, we utilize two test sets. The first set follows the official block-level evaluation protocol of Fox-Page~\citep{liu2024focus}, containing 424 text paragraph images. The second set is constructed by extracting 1,856 text paragraphs from our Dolphin-Page. Unlike page-level evaluation which considers both reading order prediction and content recognition, this element-level evaluation focuses solely on fundamental text recognition capability.

\textbf{\textit{(b) Formula.}} For formula recognition evaluation, we utilize three public benchmarks~\citep{wang2024unimernet} with different complexity levels: SPE with 6,762 simple printed expressions, SCE containing 4,742 screen capture formulas, and CPE consisting of 5,921 complex mathematical expressions. We adopt Character Difference Metric (CDM),  which measures the character-level edit distance between predictions and ground truth.

\textbf{\textit{(c) Table}.} The table recognition evaluation is conducted on two widely-used benchmarks: PubTabNet~\citep{zhong2020image} and PubTab1M~\citep{smock2022pubtables}. The test set of PubTabNet contains 7,904 table images from scientific papers, while PubTab1M's test set consists of 10,000 more challenging samples. Both benchmarks evaluate the model's capability in understanding table structures and recognizing cell contents using TEDS (Tree-Edit-Distance-based Similarity) as the metric, which computes the similarity between the predicted and ground-truth HTML table structure.

\section{Experiment}

\subsection{Implementation Details}
In the proposed Dolphin, the encoder uses a Swin Transformer with a window size of 7 and hierarchical structure ([2, 2, 14, 2] encoder layers with [4, 8, 16, 32] attention heads). The decoder contains 10 Transformer layers with a hidden dimension of 1024. We train the model using AdamW optimizer with a learning rate of 5e-5 and cosine decay schedule. The training is conducted on 40 A100 GPUs for 2 epochs, using a batch size of 16 per device through gradient accumulation.

We use normalized coordinates for bounding boxes. Specifically, we maintain the aspect ratio of input document images by first resizing the longer edge to 896 pixels, then padding to create a square image of 896$\times$896 pixels. The normalized bounding box coordinates correspond to positions within this final 896×896 padded image.

\begin{figure*}[t]
    \centering
    \includegraphics[width=2\columnwidth]{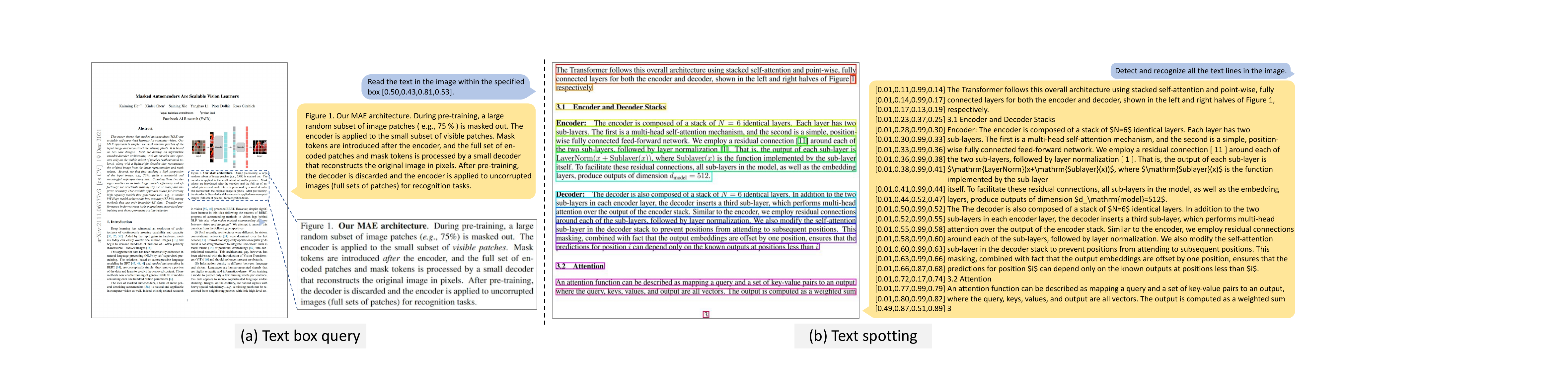}
    \caption{Additional capabilities of Dolphin. \textbf{Left}: Parsing the text content from a given bounding box region. \textbf{Right}: Text spotting results showing detected text lines (visualized in the image) and their content.}
  \label{fig:ability}
\end{figure*}

\subsection{Comparison with Existing Methods}
Comprehensive evaluations are conducted on both full-page document parsing (plain and complex documents) and individual element recognition tasks (text paragraphs, tables, and formulas).

\smallskip
\textbf{Page-level Parsing.} We evaluate Dolphin's performance on Fox-Page (English and Chinese) and Dolphin-Page benchmarks. As shown in Table~\ref{tab:benchmark-results}, despite its lightweight architecture (322M parameters), Dolphin achieves superior performance compared to both integration-based methods and larger VLMs. For pure text documents, Dolphin achieves an edit distances of 0.0114 and 0.0131 on English and Chinese test sets respectively, outperforming specialized VLMs like GOT (with edit distances of 0.035 and 0.038) and general VLMs like GPT-4.1 (with edit distances of 0.0489 and 0.2549). The advantage becomes more evident on Dolphin-Page, where Dolphin achieves an edit distance of 0.1283, outperforming all baselines in handling documents with mixed elements like tables and formulas. Furthermore, with parallel parsing design, Dolphin demonstrates considerable efficiency gains, achieving 0.1729 FPS, which is nearly 2× faster than the most efficient baseline (Mathpix at 0.0944 FPS).

We visualize three representative cases in Figure~\ref{fig:page_visualization}, showing the complete pipeline from layout analysis (Stage 1) to element-specific parsing (Stage 2), and finally to the rendered document. As demonstrated, Dolphin accurately captures both layout structure and textual content. As shown in Figure~\ref{fig:ability} (left), Dolphin also exhibits strong text extraction capabilities by accurately parsing content from specified bounding box regions.

\smallskip
\textbf{Element-level Parsing.} Beyond page-level parsing, we conduct extensive experiments to evaluate Dolphin's performance on individual elements, as shown in Table~\ref{tab:benchmark-results-block}. For text paragraph parsing, Dolphin achieves competitive results on both Fox-Block and Dolphin-Block test sets. In formula recognition, Dolphin demonstrates strong capabilities across different complexity levels (SPE, SCE, and CPE), achieving competitive CDM scores comparable to specialized formula recognition methods. For table parsing, our approach shows promising results on both PubTabNet and PubTab1M benchmarks, effectively capturing both structural relationships and cell contents. These consistent strong results across text paragraphs, formulas, and tables demonstrate Dolphin's competitive performance in fundamental recognition tasks.

We further show Dolphin's robustness in Figure~\ref{fig:network} through three scenarios: text paragraphs with complex layouts, bilingual text recognition, and structured tables with intricate formats. As shown in Figure~\ref{fig:ability} (right), Dolphin also supports text spotting by detecting and parsing text lines.

\subsection{Ablation Studies}
We conduct extensive experiments to validate the effectiveness of the core components in Dolphin.

\begin{table}[t]
\setlength{\tabcolsep}{2.5mm}
\small
\centering
\begin{tabular}{lcc}
\hline
\textbf{Method} & \textbf{ED $\downarrow$} & \textbf{FPS $\uparrow$}  \\
\hline
\textbf{Dolphin} & \textbf{0.1028} & \textbf{0.1729} \\
\hline
Parallel $\rightarrow$ Sequential Decoding & - & 0.0971 \\
Type-specific $\rightarrow$ Generic Prompts & 0.1613 & - \\
Element Cropping $\rightarrow$ Box Query & 0.1849 & - \\
\hline
\end{tabular}
\caption{\label{tab:ablation}
Ablation studies on Dolphin. The first row shows the performance of our full model. The evaluation is conducted on Dolphin-Page dataset.}
\end{table}

\smallskip
\textbf{Parallel Decoding.} To investigate the efficiency gains from our parallel decoding strategy in stage 2, we compare our approach with a sequential autoregressive decoding baseline. As present in Table~\ref{tab:ablation}, parallel decoding achieves a 1.8× speedup (0.1729 vs. 0.0971 FPS) while maintaining the same parsing accuracy. The speedup is bounded by two factors: (a) the preprocessing overhead for each element before network inference, and (b) the batch size constraint (maximum 16 elements per batch) due to GPU memory limitations, requiring multiple inference passes for documents with numerous elements. Note that existing off-the-shelf autoregressive parallel decoding solutions~\citep{kwon2023efficient} can be leveraged to further accelerate inference speed.

\begin{figure}[t]
  \centering
  \includegraphics[width=1\columnwidth]{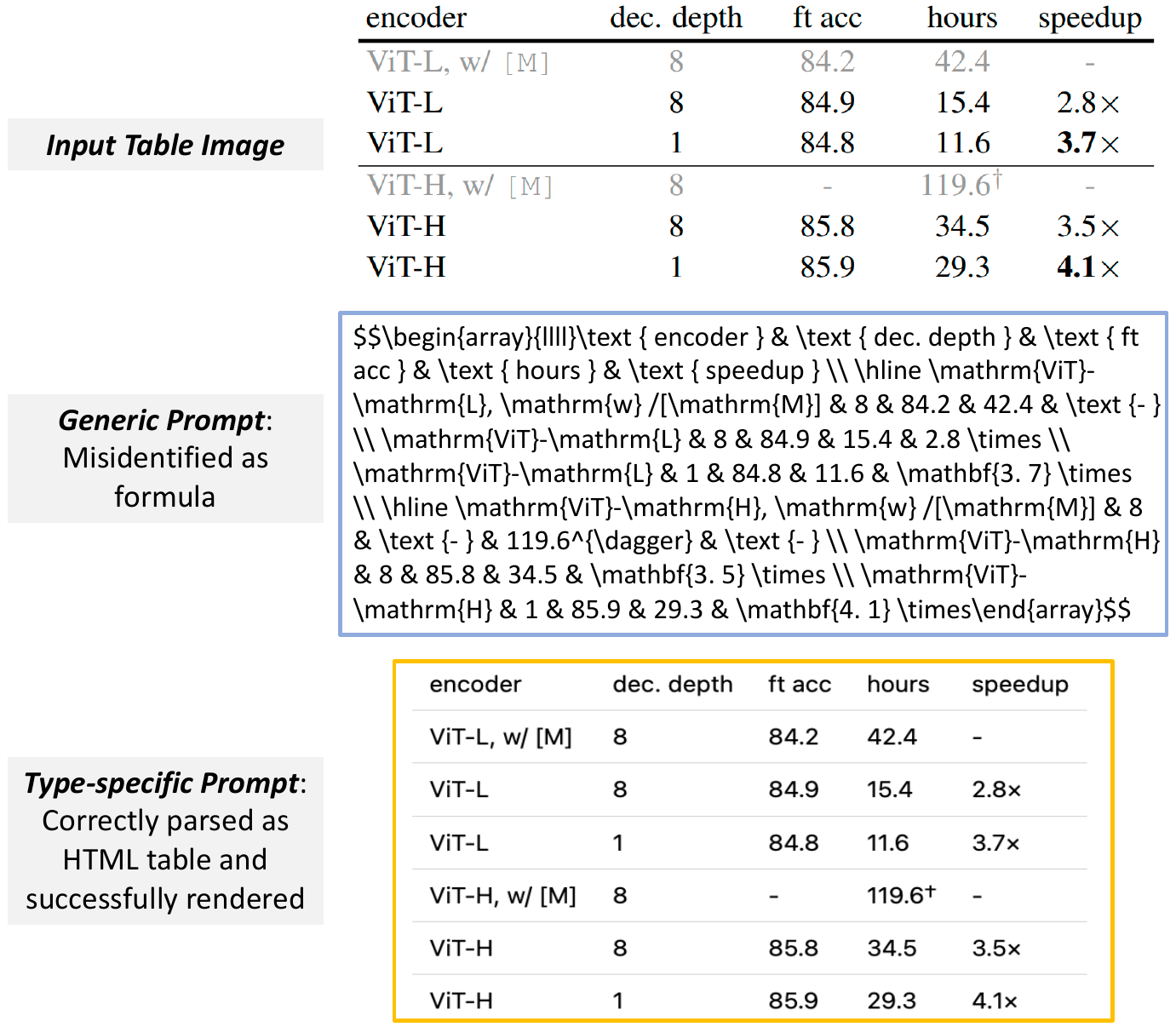}
    \caption{A case study demonstrating the effectiveness of type-specific prompts. The generic prompt misidentifies the table as a formula, while our approach correctly parses and renders the table in HTML format.}
  \label{fig:prompt_aba}
\end{figure}

\smallskip
\textbf{Type-specific vs. Generic Prompts.} To investigate the effectiveness of type-specific prompting in the second stage, we compare Dolphin with a baseline variant that uses a generic prompt "\textit{Read text in the image.}" for all element parsing tasks. As shown in Table~\ref{tab:ablation}, our type-specific prompting strategy significantly outperforms the generic baseline (0.1283 vs. 0.1613 in ED). A representative case is shown in Figure~\ref{fig:prompt_aba}, where the generic prompt misidentifies a table as a LaTeX formula, while our type-specific prompt successfully parses and renders it. These results demonstrate that incorporating prior knowledge through type-specific prompting effectively improves the model's ability to handle different document elements.

\smallskip
\textbf{Element Cropping vs. Box Query.} To validate our element cropping strategy in the second stage, we compare it with an alternative box query approach that directly prompts the model to recognize elements at specific box (see Figure~\ref{fig:ability} (left)). As shown in Table~\ref{tab:ablation}, our cropping strategy achieves better performance than the box query method. This is likely because cropping provides the model with a focused view of each element, following a ``what you see is what you get'' principle, while the box query approach increases task complexity by requiring the model to simultaneously handle location understanding and content recognition.

\section{Conclusion}
We present Dolphin, a novel document image parsing model that leverages an analyze-then-parse paradigm to address the challenges in document parsing. Our approach first performs page-level layout analysis to generate structured layout elements in reading order, then enables parallel element parsing through heterogeneous anchor prompting. This two-stage design effectively balances efficiency and accuracy, while maintaining a lightweight architecture. Through extensive experiments, we demonstrate Dolphin's strong performance in both page-level and element-level parsing tasks, particularly excelling in handling complex documents with interleaved tables, formulas, and rich formatting in both Chinese and English.

\section*{Limitations}
Despite Dolphin's promising performance, there are several limitations worth noting. First, Dolphin primarily supports documents with standard horizontal text layout, showing limited capability in parsing vertical text like ancient manuscripts. Second, while Dolphin handles both Chinese and English documents effectively, its multilingual capacity~\citep{tang2024mtvqa} needs to be expanded. Nevertheless, we demonstrate some cases exhibiting emergent multilingual document parsing capabilities in the supplementary materials. Third, although we achieve efficiency gains through parallel element parsing, there is potential for further optimization through parallel processing of text lines and table cells. Fourth, handwriting recognition capabilities require further enhancement.

\bibliography{main}

\appendix
\newpage
\twocolumn

In this supplementary material, we provide additional experimental results and implementation details to 
complement our main paper. Specifically, we present more qualitative results demonstrating Dolphin's parsing capabilities, elaborate on the supported element types, detail our training process, and showcase our synthetic data.

\section{Qualitative Results}
To further demonstrate the superior capabilities of Dolphin, we present comprehensive page-level and element-level parsing results.

\begin{figure*}[t!]
    \centering
    \includegraphics[width=1\textwidth]{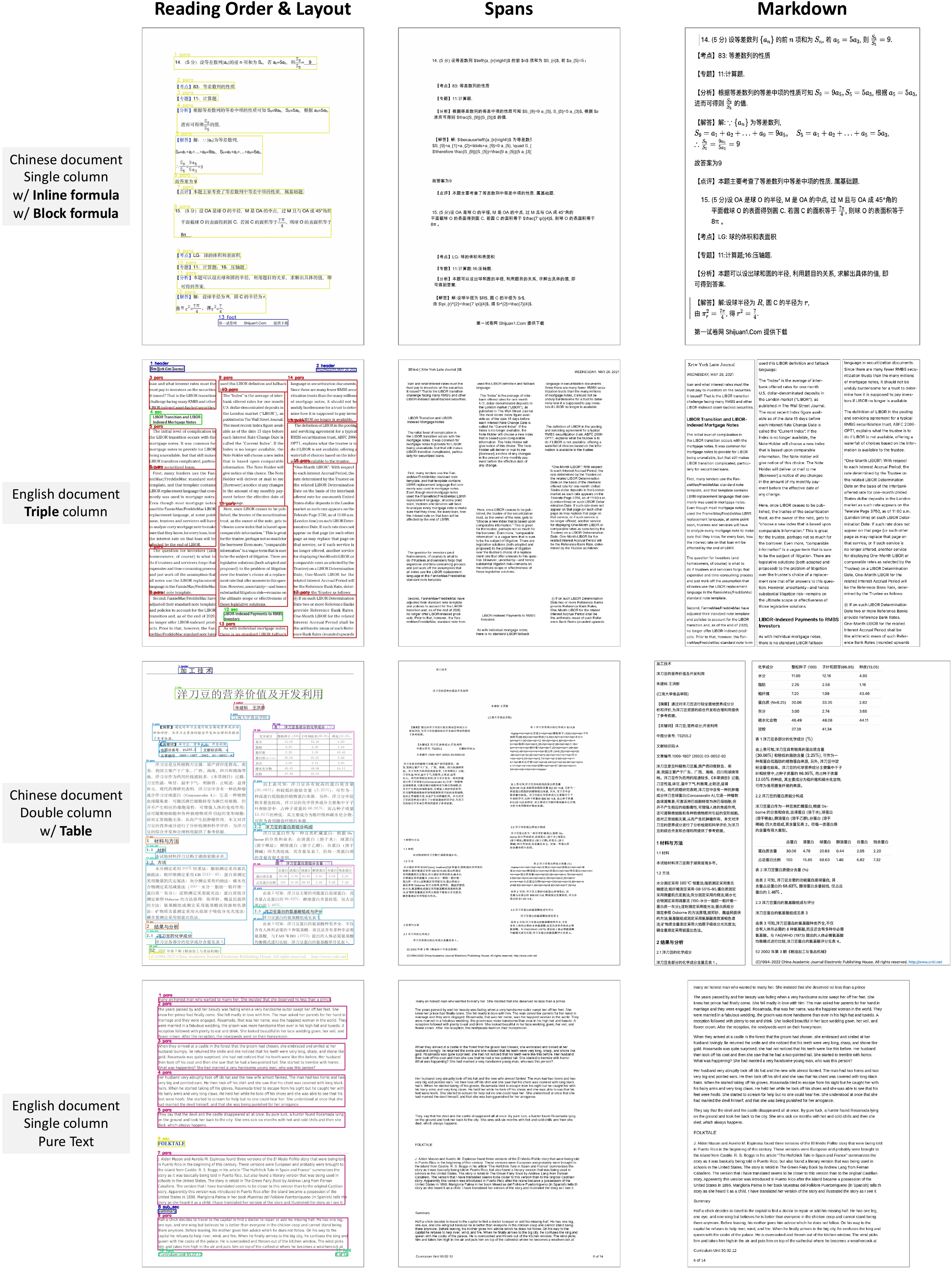}
    \caption{Visualization of Dolphin's \textbf{page-level} parsing results. \textbf{Left}: Layout analysis form Stage 1 with predicted element boundaries and reading order. \textbf{Middle}: Element-specific parsing outputs from Stage 2. \textbf{Right}: Final rendered document in markdown format.}
    \label{fig:supp_page_visualization_1}
\end{figure*}

\textbf{Page-level.} First, the examples in Figure~\ref{fig:supp_page_visualization_1} cover diverse document scenarios, including textbook pages with dense formulas, triple-column English academic papers, and double-column Chinese papers with tables. The results demonstrate that Dolphin can effectively handle documents with different languages, layouts, and element types, maintaining high parsing quality.

\begin{figure*}[t!]
    \centering
    \includegraphics[width=1\textwidth]{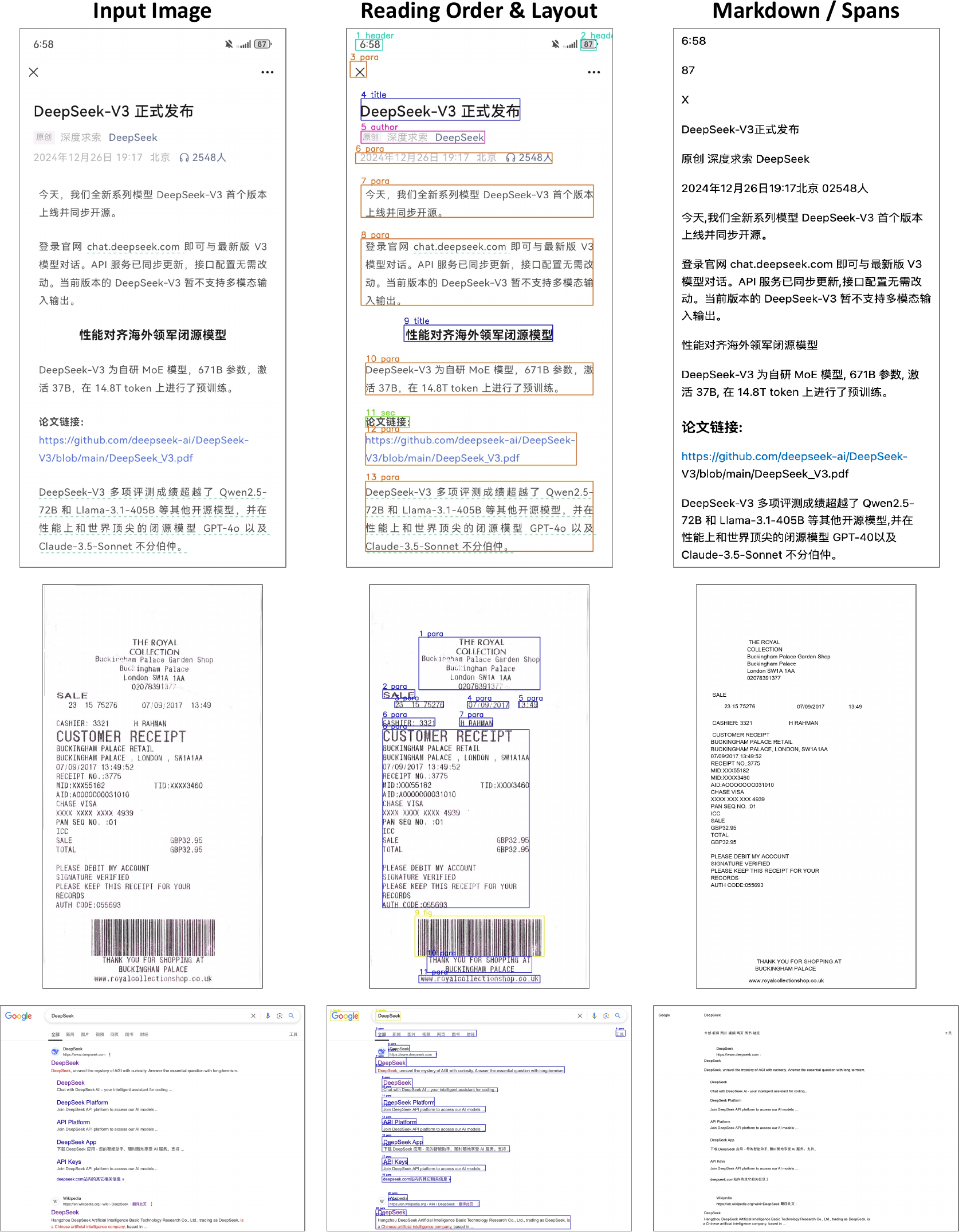}
    \caption{Visualization of Dolphin's \textbf{page-level} parsing results. \textbf{Left}: Input text-rich images including mobile phone screenshots, shopping receipts, and webpage captures. \textbf{Middle}: Layout analysis form Stage 1 with predicted element boundaries and reading order. \textbf{Right}: Final rendered document in markdown format for the first row, and element-specific parsing outputs from Stage 2 for the second and third rows.}
    \label{fig:supp_page_visualization_3}
\end{figure*}

Furthermore, we showcase Dolphin's versatility in other text-rich scenarios through Figure~\ref{fig:supp_page_visualization_3}, where we test the model on mobile phone screenshots, shopping receipts, and webpage captures. These results indicate that Dolphin can accurately capture both the structural layout and textual content in these everyday scenarios.

\begin{figure*}[t!]
    \centering
    \includegraphics[width=1\textwidth]{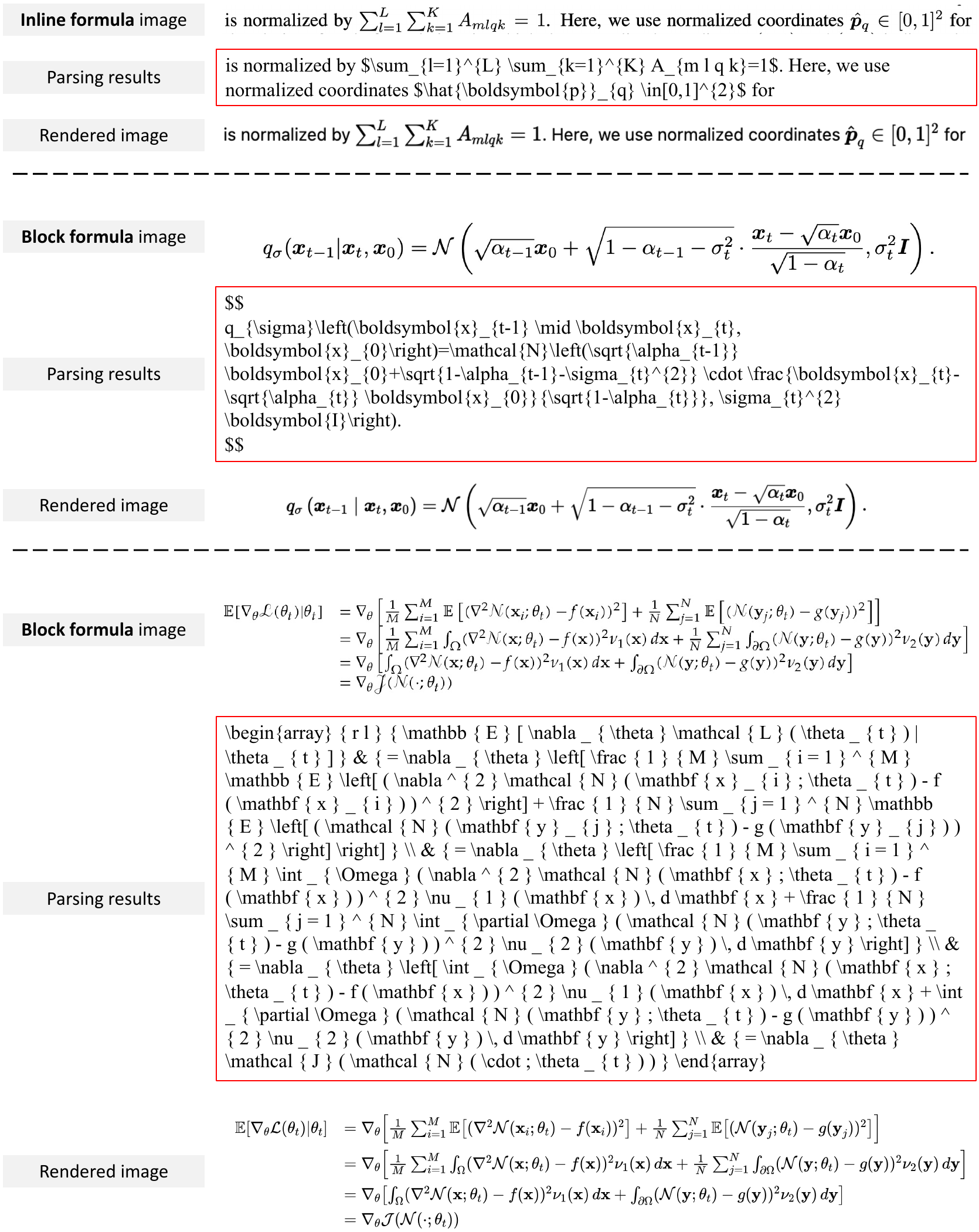}
    \caption{Visualization of Dolphin's \textbf{formula} parsing results. From top to bottom, we show three formula types: \textbf{inline formula}, \textbf{single-line block formula}, and \textbf{multi-line block formula}. For each case, we visualize the complete parsing pipeline: input formula image (top), LaTeX parsing output (middle), and rendered formula (bottom). These results demonstrate Dolphin's capability to accurately parse formulas of varying complexity.}
    \label{fig:supp_formula}
\end{figure*}

\textbf{Element-level.} For fine-grained parsing capabilities, we first demonstrate Dolphin's formula recognition in Figure~\ref{fig:supp_formula}, where we evaluate three types of formulas: inline formulas, single-line block formulas, and multi-line block formulas. The results show that Dolphin can accurately parse formulas of varying complexity and layout formats.

\begin{figure*}[t!]
    \centering
    \includegraphics[width=0.75\textwidth]{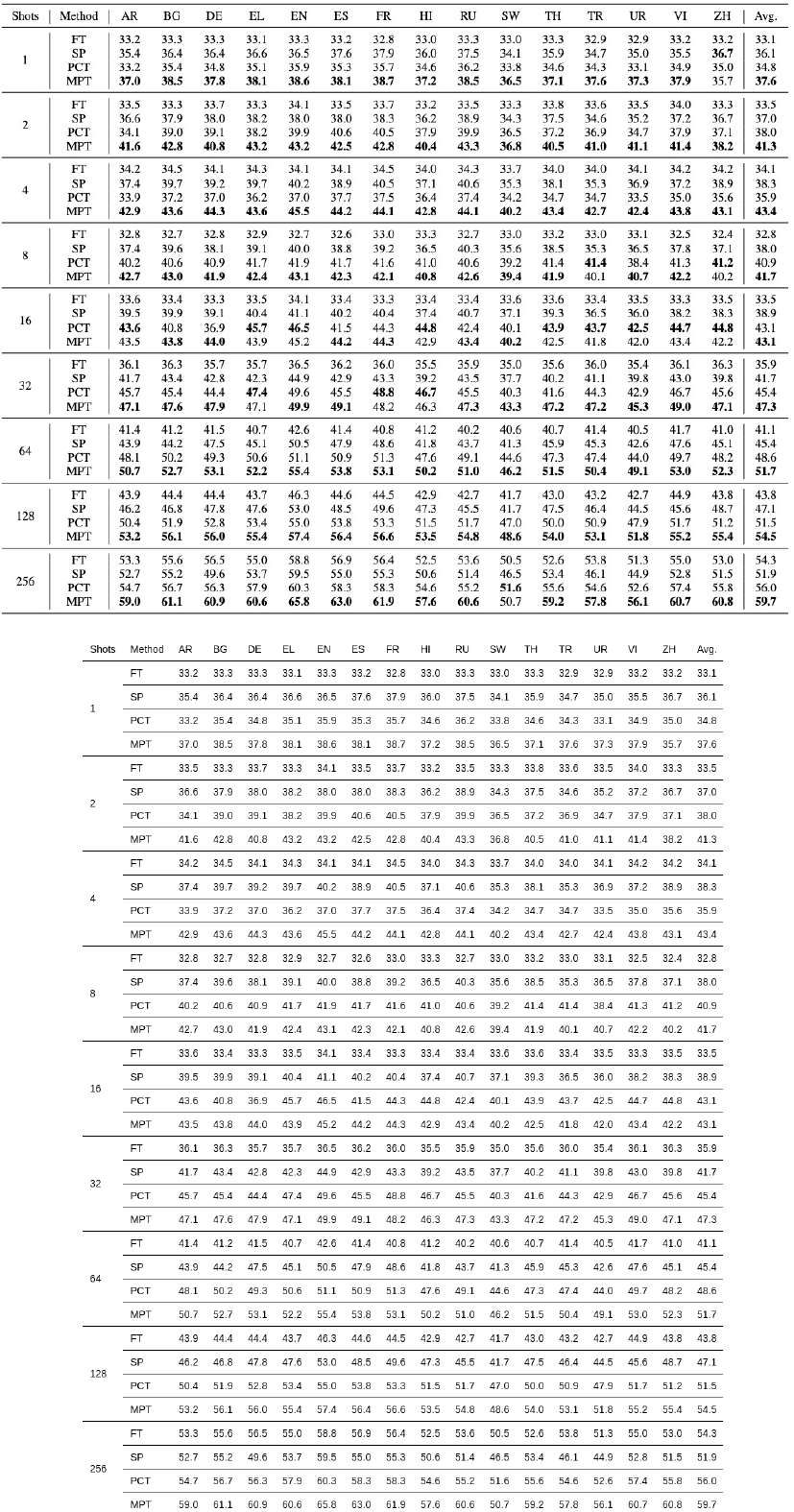}
    \caption{Visualization of Dolphin's \textbf{table} parsing capability. \textbf{Top}: Input large-scale table image containing hundreds of cells. \textbf{Bottom}: Rendered HTML table based on Dolphin's parsing result. This example demonstrates Dolphin's strong ability in handling large-scale structured table images.}
    \label{fig:supp_table}
\end{figure*}

We further evaluate Dolphin's table parsing ability in Figure~\ref{fig:supp_table}, where we test the model on a challenging case containing hundreds of cells. As shown, Dolphin successfully handles this large-scale structured table with precise content recognition and layout preservation.

\section{Element Design}
In this section, we elaborate on Dolphin's supported element types and element-specific parsing strategies through heterogeneous prompting.

\textbf{Element Types.}
Our Dolphin supports 15 different types of elements commonly found in document images. Table~\ref{tab:element_types} provides a comprehensive overview of these elements, covering various components from headers to specialized content blocks. 

Note that in Stage 1 (page-level layout analysis), we intentionally avoid treating formulas as independent elements. This design choice allows Stage 2 (element-level parsing) to leverage broader contextual information when recognizing mathematical expressions, as formulas are often semantically connected with their surrounding text. 

\textbf{Heterogeneous Anchor Prompting.}
We summarize the prompts used in Dolphin in Table~\ref{tab:prompt_types}. The first three prompts (page-level layout analysis, text paragraph parsing, and table parsing) are designed for full-page document image parsing, while the latter two (text spotting and text box query) enable additional capabilities for flexible text recognition tasks. Additionally, our
Dolphin can also serve as a formula recognition expert model using the text paragraph parsing prompt.

In Stage 2, tables are processed with a dedicated table-specific prompt for structured HTML parsing, while all other elements are treated as text paragraphs and parsed using a unified prompt. This dichotomous design distinguishes structured HTML content from plain text, while also providing robustness against potential element misclassification, as parsing accuracy remains high regardless of element type classification errors. 

\section{Training Details}
In this section, we provide more details about Dolphin's training process, including multi-task training strategy, model initialization, and other implementation considerations.

\begin{table}[t]
\setlength{\tabcolsep}{2.6mm}
\small
\centering
\begin{tabular}{ccl}
\toprule
\textbf{No.} & \textbf{Element} & \textbf{Description} \\
\midrule
1 & title & Paper/document title \\
2 & author & Author names \\
3 & sec & First-level section headings \\
4 & sub\_sec & Second-level section headings \\
5 & para & Paragraphs \\
6 & header & Page headers \\
7 & foot & Page footers \\
8 & fnote & Footnotes \\
9 & watermark & Non-content watermarks \\
10 & fig & Figures and images \\
11 & tab & Tables \\
12 & cap & Figure/table captions \\
13 & anno & Figure/table annotations \\
14 & alg & Code blocks/pseudocode \\
15 & list & List-type content \\
\bottomrule
\end{tabular}
\caption{An overview of element types supported by Dolphin. These elements cover the majority of content structures found in documents.}
\label{tab:element_types}
\end{table}

\begin{figure*}[t]
    \centering
  \includegraphics[width=1.87\columnwidth]{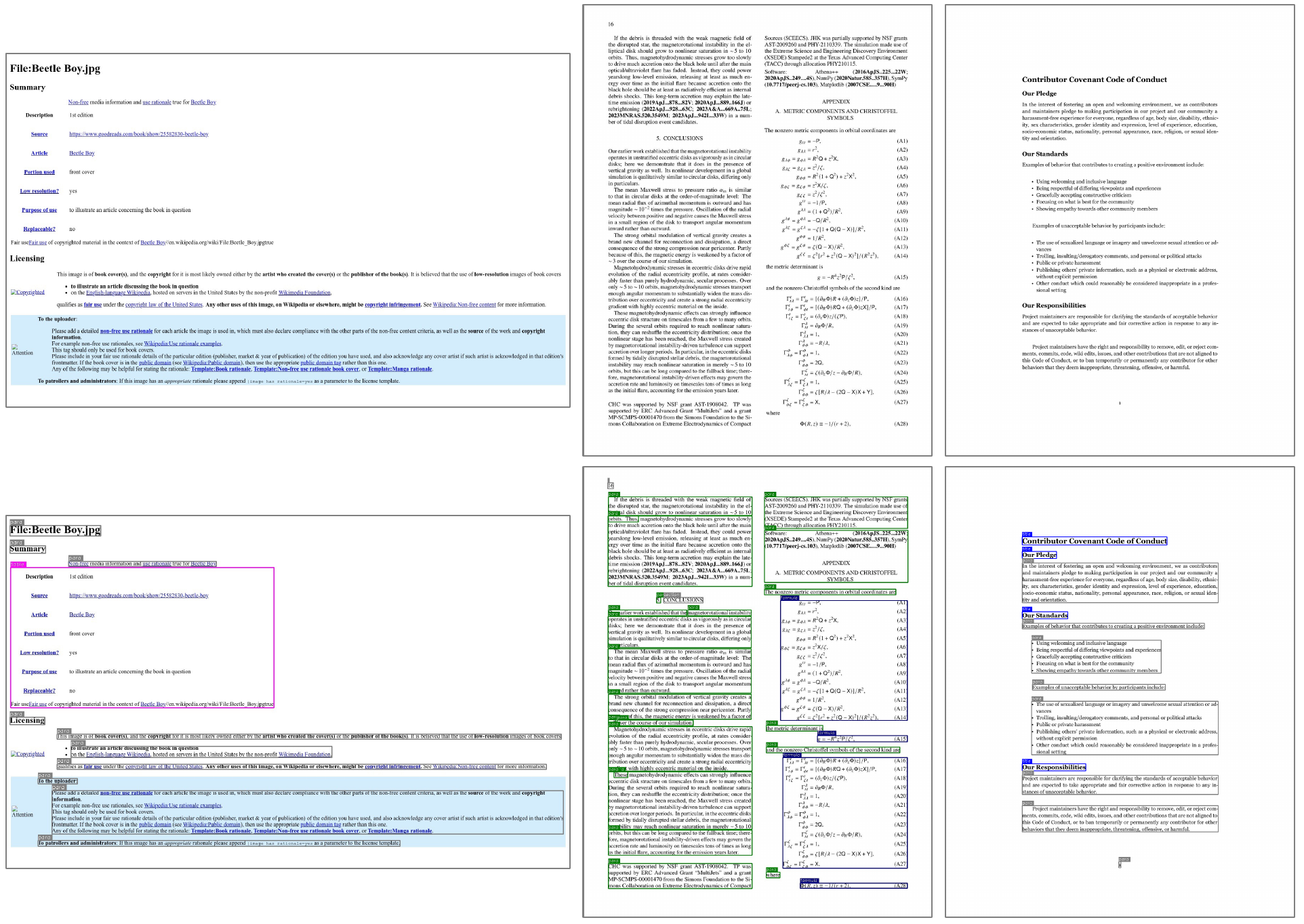}
    \caption{Examples of synthetic training data generated from different source formats. \textbf{Top}: rendered document images from HTML (left), LaTeX (middle), and Markdown (right) sources. \textbf{Bottom}: corresponding paragraph-level annotations visualized with colored regions.}
  \label{fig:data_case}
\end{figure*}

\begin{table*}[t]
\setlength{\tabcolsep}{3mm}
\small
\centering
\begin{tabular}{c|c}
\toprule
\textbf{Task} &  \textbf{Prompt} \\
\midrule
Page-level Layout Analysis & Parse the reading order of this document. \\
\midrule
Text Paragraph/Formula Parsing & Read text in the image. \\
\midrule
Table Parsing & Parse the table in the image. \\
\midrule
Text Spotting & Detect and recognize all the text lines in the image. \\
\midrule
Text Box Query & Read the text in the image within the specified box [x1,y1,x2,y2]. \\
\bottomrule
\end{tabular}
\caption{Different types of prompts used in Dolphin for document parsing tasks.}
\label{tab:prompt_types}
\end{table*}

\textbf{Instruction Tuning.}
During training phase, we adopt a dynamic task selection strategy for our instruction-based framework. Specifically, given a training sample, we randomly select an applicable task from the above five tasks based on its available annotations. This selection is used to construct question-answer pairs. For instance, given a page image with only paragraph-level bounding boxes and content annotations, the available tasks for this sample would include element-level text paragraph parsing and page-level box query parsing.

\textbf{Model Initialization.}
We initialize Dolphin with the pretrained weights from Donut~\citep{kim2022ocr}, which lacks instruction-following abilities. Then, through our instruction tuning, we extend the model's capabilities to understand and execute diverse prompts, enabling analysis of document layout, reading order, and various textual elements including text paragraphs, tables, and formulas.

\textbf{Training Loss.}
Following standard practice in autoregressive language models, we optimize Dolphin using the cross-entropy loss between the predicted token distributions and ground truth ones. 

\section{Synthetic Data Examples}
To enrich training data diversity, we synthesize document images from different source formats, including HTML, LaTeX, and Markdown documents. Figure~\ref{fig:data_case} shows three representative examples of our synthetic data. For each format, we show the rendered document (top row) and its corresponding paragraph-level annotations (bottom row).

\end{document}